\documentclass[pmlr]{jmlr} 

\jmlrvolume{}
\jmlrproceedings{}{}

\usepackage{booktabs} 
\usepackage{multirow} 
\usepackage{graphicx} 
\usepackage{url}      
\usepackage{amsmath}  
\usepackage{algorithm2e} 
\RestyleAlgo{ruled}
\usepackage{dirtytalk} 
\usepackage[normalem]{ulem} 

\usepackage{soul} 
\definecolor{lightblue}{RGB}{198, 226, 255}  

\usepackage[textsize=tiny,textwidth=1.8cm]{todonotes}  
\newcommand{\change}[1]{\todo[color=yellow!80, bordercolor=black]{#1}}

\newcommand{\deleted}[1]{}  
\DeclareRobustCommand{\hl}[1]{#1} 
\renewcommand{\todo}[2][]{}

\usepackage{float} 
\usepackage[T1]{fontenc}

\jmlrvolume{}
\jmlryear{}
\jmlrpages{}
\jmlrworkshop{Accepted as a poster ate Probabilistic Graphical Models (PGM) 2026}

\title{LLM-Augmented Causal Discovery: Probabilistic Fusion of Edge Existence and Orientation}

\author{\Name{Neville K. Kitson} \Email{n.k.kitson@causaliq.org\thanks{Alternative email address: n.k.kitson@qmul.ac.uk}}\\
  \AND
  \Name{Anthony C. Constantinou} \Email{a.constantinou@qmul.ac.uk}\\
  \addr \\
  Machine Intelligence and Decision Systems (MInDS) Group, 
  School of Electronic Engineering and Computer Science, Queen Mary University of London, London, United Kingdom, E1 4NS.
}

\editor{Gustau Camps-Valls, Manuele Leonelli and Gherardo Varando}

\begin{document}

\maketitle

\begin{abstract}
Bayesian network structure learning (BNSL) from observational data struggles with orientation identifiability, while large language models (LLMs) offer broad but often unreliable causal knowledge. We propose combining these complementary sources through a novel representation, termed Probabilistic Dependency Graphs (PDGs). In a PDG, each edge is associated with a distribution over directed, undirected, and absent states, enabling fusion via weighted averaging. We evaluate this approach on 26 benchmark networks, combining ensembles of three BNSL algorithms (FGES, Tabu, PC) with three LLMs (Gemini, Claude, GPT) across multiple prompts and random seeds. A simple 50/50 fusion improves F1 over the better of either source alone in 22 of 26 networks, with a statistically significant mean improvement of $0.056$ $(p<0.001)$. Analysis reveals that the two sources play complementary roles: BNSL contributes a high-recall edge skeleton (80\% vs 60\% for LLM), while LLM contributes accurate edge orientation (96\% vs 77\% for BNSL). Our results show that representing both sources as probabilistic uncertainty over edge existence and orientation is a practical and effective way to improve causal graph accuracy.
\end{abstract}


\section{Introduction}

\setlength{\marginparwidth}{2cm} 

Bayesian networks (BNs) compactly encode the joint probability distribution over a set of variables as a directed acyclic graph (DAG) together with parameters defining the strength of dependencies \citep{koller2009probabilistic}.  Two broad families of algorithms have been developed for BN structure learning (BNSL). Constraint-based methods such as PC \citep{spirtes2000causation} use conditional independence tests to identify the skeleton and orientate edges, returning a completed partially directed acyclic graph (CPDAG) that represents the Markov equivalence class: the set of DAGs encoding the same conditional independencies. Score-based methods such as GES \citep{chickering2002optimal} and greedy hill-climbing algorithms search the space of equivalence classes or DAGs by optimising a penalised likelihood score such as BIC or BDeu. BNSL remains challenging in practice. Finite-sample data yields unreliable independence tests and noisy score landscapes, whilst equivalence classes can be large, leaving many edges undirected in CPDAGs, or arbitrarily orientated arcs in DAGs. Assumptions of no noise or latent variables, inappropriate probability distributions and functional relationships also contribute to inaccuracy \citep{kitson2023survey}. 

These limitations have motivated a growing interest in incorporating prior knowledge to guide or constrain BNSL. Large language models (LLMs) represent a new and readily available source of such knowledge\deleted{. Recent work has explored using LLMs in conjunction with BNSL Ban et al., 2025; Bazaluk et al., 2025; Constantinou et al., 2025; Vashishtha et al., 2025), typically by querying pairwise edge existence or using LLM outputs as structural priors. However}\hl{, although}\change{\textbf{rev2:} provide more details of other approaches for combining BNSL and LLMs, and contrast with our proposed approach} LLM-proposed graphs tend to be imprecise; they may hallucinate edges, miss domain-specific relationships, or reflect correlational rather than causal patterns \citep{liu2025large, wan2025large}. \hl{Recent work has combined LLMs with causal discovery in several ways. }\citet{constantinou2025using}\hl{ employ GPT-4 to define constraints on direct edges, }\citet{ban2025integratingllm}\hl{ use LLM-derived ancestral constraints, }\citet{vashishtha2025causal}\hl{ use LLM-derived causal ordering information, and }\citet{bazaluk2025large}\hl{ employ LLMs to define probabilistic priors. In contrast, our objective is not to replace or constrain structure learning, but to represent and fuse uncertainty from statistical and LLM-derived sources via a common probabilistic representation.}

This study tests the hypothesis that the errors made by BNSL and LLMs are complementary and that combining the two in a way that retains the strengths of each may lead to more accurate causal discovery. We introduce Probabilistic Dependency Graphs (PDGs), a new representation of causal graph structure which separately captures existence and orientation uncertainty and supports a principled fusion of the two sources. While related to model averaging in structure learning, PDGs differ in that they represent uncertainty directly at the edge level rather than as weights over complete graph models.

The remainder of the paper is structured as follows. Section~\ref{sec:methods} introduces PDGs and describes how LLM‑derived and data‑driven structures are represented and fused. Section~\ref{sec:results} presents an experimental evaluation on benchmark networks and analyses the complementary roles of data‑driven and LLM‑based sources. Section~\ref{sec:conclusion} discusses limitations and directions for future work.

\section{Methods}
\label{sec:methods}

This section presents our approach for integrating data‑driven and knowledge‑based causal information via a unified probabilistic representation. We introduce PDGs to capture uncertainty in both edge existence and orientation, and show how outputs from BNSL and LLM‑derived causal knowledge are mapped to this representation and fused.

\subsection{Probabilistic Dependency Graphs}
\label{sub:pdg}

We introduce the concept of a PDG as a means of capturing uncertainty in edge existence and edge
orientation. Edges are defined by a tuple of the two endpoints, $(A, B)$ where $A < B$ lexicographically, and four
probabilities which sum to 1.0:\change{\textbf{pc1:} change p subscripts to match those in algorithm pseudo-code, and change to inline text to save space} $p_{\text{none}}$\hl{ for no edge; }$p_{\text{fwd}}$\hl{ for directed edge }$A \rightarrow B$\hl{; }$p_{\text{bwd}}$\hl{ for directed edge }$A \leftarrow B$;\hl{ and }$p_{\text{und}}$\hl{ for an undirected edge}. This formulation means that PDGs can represent conventional DAGs and CPDAGs using $p_{\text{fwd}} = 1.0$ or $p_{\text{und}} = 1.0$ for directed or undirected edges respectively. However, when non-unit probabilities are used,
PDGs capture edge existence and orientation uncertainty.

\SetAlgoNoEnd
\vspace{6pt}
\begin{algorithm2e}[H]
\DontPrintSemicolon
\SetAlgoLined
\KwIn{Graphs $G_1, \ldots, G_n$ 
      each with nodes $\mathbf{V}$;
      optional weights $w_1, \ldots, w_n$, $\sum_i w_i = 1$}
\KwOut{Fused PDG $G^{*}$ over $\mathbf{V}$}
\medskip
\If{weights not provided}{
    $w_i \leftarrow 1/n$ for all $i$\;
}
\ForEach{graph $G_i$}{
    \If{$G_i$ is a DAG, CPDAG or PDAG}{
        Convert $G_i$ to a PDG by setting
        $p_{\text{forward}} = 1$ for directed edges,
        $p_{\text{undirected}} = 1$ for undirected edges,
        and $p_{\text{none}} = 1$ for absent edges\;
    }
}
Initialise $G^{*}$ as a PDG over $\mathbf{V}$ with no edges\;
\ForEach{canonical node pair $(A, B)$ with $A < B$}{
        Let $\mathbf{p^{*}} = (p^{*}_{\text{fwd}},\, p^{*}_{\text{bwd}},\,
              p^{*}_{\text{und}},\, p^{*}_{\text{none}}) = (0, 0, 0, 0)$
        be the probabilities for $(A, B)$ in $G^{*}$\;
    \For{$i \leftarrow 1$ \KwTo $n$}{
        $\mathbf{p^{*}} \leftarrow \mathbf{p^{*}} + w_i \cdot \mathbf{p}^i$\;
    }
    \If{$p^{*}_{\text{fwd}} + p^{*}_{\text{bwd}}
         + p^{*}_{\text{und}} > 0$}{
        Set probabilities of $(A,B)$ in $G^{*}$ to
        $(p^{*}_{\text{fwd}},\, p^{*}_{\text{bwd}},\,
          p^{*}_{\text{und}},\, p^{*}_{\text{none}})$\;
    }
}
\Return{$G^{*}$}\;
\caption{\textsc{MergeGraphs} --  fuse graphs into a PDG.}
\label{alg:merge_graphs}
\end{algorithm2e}
\vspace{6pt}

\hl{Algorithm~}\ref{alg:merge_graphs}\hl{ (\textsc{MergeGraphs}) is used to merge an arbitrary number of graphs, with an optional weighting assigned to each one. The graphs can be any mixture of DAGs, CPDAGs, non-extendable PDAGs and PDGs.}\change{\textbf{pc1: }algorithms \textsc{MergeGraphs} and \textsc{ToDagGreedy} and accompanying descriptive text moved here from Appendices, and pseudo-code defined more concisely} This algorithm is used in a variety of situations in this study, for example, to "average" graphs produced by multiple runs of BNSL algorithms, to build ensemble PDGs across multiple BNSL algorithms or LLM models, and to fuse the final LLM and BNSL PDGs. \hl{PDG values should be interpreted as empirical support for edge states rather than calibrated posterior probabilities. For LLMs these values originate from elicited confidence scores, whilst for BNSL they reflect frequencies across ensembles of learnt graphs. }\change{\textbf{rev5:} Clarify that PDG edge probabilities do not represent calibrated posterior probabilities.}\hl{Note that PDGs do not induce variability where none exists. If all constituent sources agree on an edge, the fused PDG approaches a deterministic representation. Variability therefore reflects disagreement between sources rather than being introduced by the fusion process itself.}\change{\textbf{rev4:} Clarify that merging graphs does not introduce uncertainty where the merged graphs agree.} 

\begin{algorithm2e}[h!t]
\DontPrintSemicolon
\KwIn{PDG $G$ over node set $\mathbf{V}$; existence threshold $\tau$}
\KwOut{DAG $D$ over $\mathbf{V}$}
\medskip
$\mathcal{C} \leftarrow \emptyset$; $D \leftarrow$ empty DAG; \tcp*{$\mathcal{C}$ = candidate edge list}
\ForEach{edge $(A, B)$ in $G$}{
    \lIf{$p_{\mathrm{fwd}} + p_{\mathrm{bwd}} + p_{\mathrm{und}} < \tau$}{\textbf{skip}}
    \uIf{$p_{\mathrm{fwd}} \geq p_{\mathrm{bwd}}$}{
        add $(p_{\mathrm{fwd}} + 0.5 \times p_{\mathrm{und}},\; A \to B)$ to $\mathcal{C}$\;
    }
    \uElse{
        add $(p_{\mathrm{bwd}} + 0.5 \times p_{\mathrm{und}},\; B \to A)$ to $\mathcal{C}$\;
    }
}
\ForEach{$(p,\; X \to Y)$ in $\mathcal{C}$ in descending order of $p$}{
    \lIf{$X \to Y$ creates cycle}{\textbf{skip}}
    add $X \to Y$ to $D$\;
}
\Return{$D$}\;
\caption{\textsc{ToDagGreedy} -- extract a DAG
    from a PDG by greedy edge inclusion.}
\label{alg:to_dag_greedy}
\end{algorithm2e}

For evaluation, a directed acyclic graph is extracted from each PDG using \hl{Algorithm~}\ref{alg:to_dag_greedy}\hl{ (\textsc{ToDagGreedy})}. It is a simple greedy \hl{heuristic} procedure\hl{ rather than an optimal MAP estimator over DAG space}.\change{\textbf{rev5}: acknowledge that this is a simple heuristic approach and that more sophisticated approaches are possible.}\hl{ The algorithm first compiles a list of edges which have an existence probability }$(1-p_{none})$\hl{ greater than a threshold value, }$\tau$\hl{ , set to 0.3 in these experiments. The most likely direction for the edge is determined by comparing }$p_{fwd}$\hl{ to }$p_{bwd}$\hl{ with ties being broken lexicographically. The algorithm adds these directed arcs to the output DAG in decreasing order of their existence probability, but ignoring any arcs that would create a cycle.}

\reversemarginpar
\setlength{\marginparwidth}{2cm} 

\subsection{Generating the LLM Ensemble PDG}

\begin{figure}[!ht]
\centering
\includegraphics[width=1.0\textwidth]{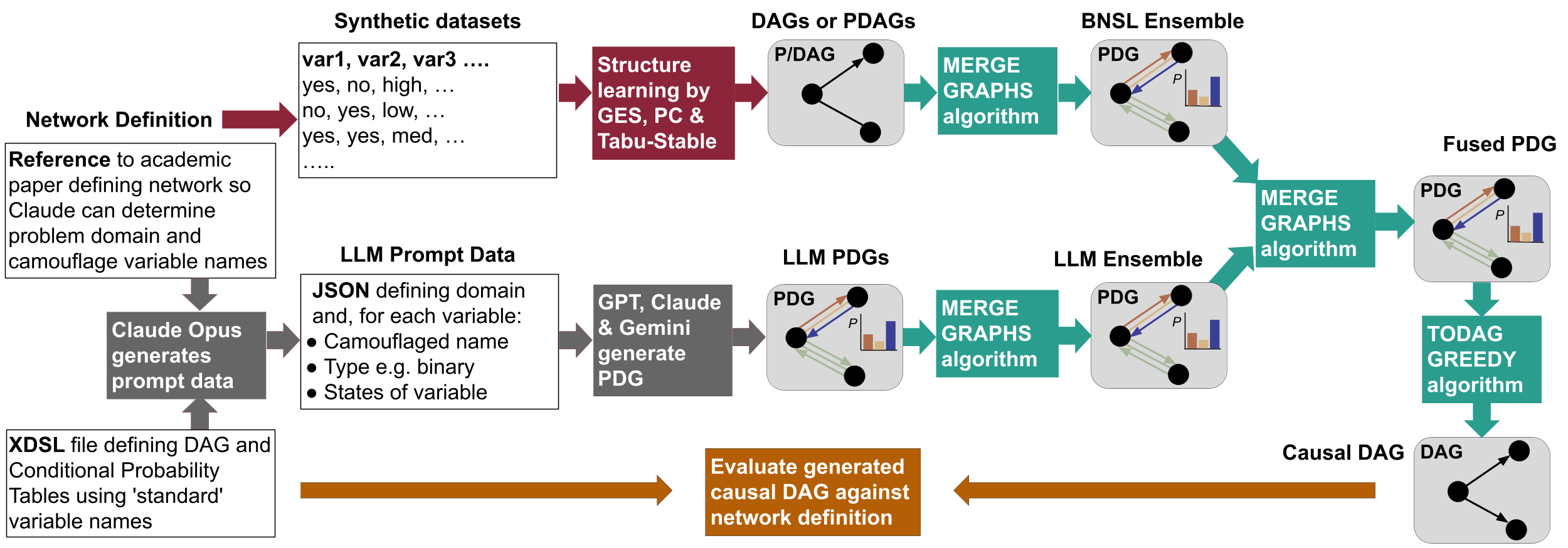}
\caption{\deleted{Overall methodology for c}\hl{C}reating a causal DAG by fusing LLM knowledge and statistical learning. Maroon-coloured flows represent standard BNSL, grey flows are LLM interactions, and teal-coloured flows show the fusion of BNSL and LLM outputs to arrive at a final causal DAG. This is then evaluated against the original network definition.}
\label{fig:pdg_merge}
\end{figure}

Figure~\ref{fig:pdg_merge}\change{\textbf{pc1:} all text in figure increased in size to improve legibility.} shows the overall methodology for fusing LLM and BNSL information to create a causal DAG for a given network; the grey boxes show interactions with LLMs. To mitigate the risk of LLMs reproducing memorised benchmark structures, variable names are replaced with neutral aliases before querying the LLM. For example, in the well-known Asia network, the variable name "asia" is presented as "endemic\_travel", and the network domain is specified as "respiratory\_diagnosis" rather than "Asia".

The network domains, neutral variable aliases, descriptions, types and states used in all causal elucidation requests are generated as a single preprocessing request to Claude Opus 4.6 LLM. This step is part of the experimental setup rather than the causal discovery method itself and is restricted to domain‑level information that a practitioner reasonably familiar with the application domain would be expected to have\deleted{; it does not encode causal relationships or structural hints}\hl{, although some causal implications may naturally arise from the semantics of the generated descriptions}\change{\textbf{pc2:} Leakage audit showed that some causal hints are presented in the generated names and descriptions}. To avoid introducing human bias and to ensure consistency across networks, this step is delegated entirely to an LLM and executed once per network using the Claude user interface. Claude is provided with the benchmark BN definition\footnote{The BN definition is provided in XML format for Directed Systems Languages (XDSL) which defines the DAG structure and Conditional Probability Tables (CPTs)}, a reference to the original academic paper, and an example JSON format, and is asked to generate a corresponding JSON file containing these details for the target network.

These LLM-generated "camouflaged" details are used in all subsequent requests to LLMs to elucidate causal relationships as follows. Each LLM receives a system prompt defining its role as a causal reasoning expert and specifying the required JSON response format: an array of edges with existence probability and orientation confidence which can then be encoded directly as a PDG. An example request for edge probabilities for the Asia network is shown in Appendix~\ref{apdx:llm_prompt}. The user prompt lists the network's variables at one of two detail levels: minimal (variable names only) or standard (names, types, descriptions, and discrete states). In order to limit response size, the system prompt includes a stipulation that the LLM should only return edges which it believes have a probability of existing greater than 0.5.

Three LLM models are used to generate the LLM PDGs in this study: Anthropic's claude-haiku-4-5-20251001, Google's gemini-2.5-flash and OpenAI's gpt-4.1-mini. They were selected on the basis of being capable enough to suggest causal relationships and reliably return a valid JSON response, but not so expensive as to discourage extensive use. Requests are made completely automatically using each vendor's published API. \deleted{The first two models are free, and the latter is low-cost, with requests costing on average 0.16 US cents in these experiments. Each of these models places a limit on the number of response tokens of 16K, 64K and 32K respectively. Table~4 in Appendix~A provides details of the number of tokens in requests and responses, as well as cost and latency. The number of response tokens for the largest networks is no more than one third of any model's limit indicating that there is some headroom to extend this approach to larger networks. Response latency is considerable, with the slowest model, gemini, taking over two minutes to provide a response for the largest networks.}\hl{ Further details of token usage, latency and API costs are provided in Appendix~}\ref{apdx:llm_prompt}\hl{.}\change{These API details moved to Appendix A to save space}

As we will see in the results, maintaining diversity in the generated PDGs contributes considerably to the quality of the final result. Each request for a given network at either minimal or standard prompt detail is sent to each LLM model three times. The LLM temperature hyperparameter controls the randomness of its responses; a relatively high value of 0.7 is used to encourage different responses, including from the trios of identical requests. A final ensemble LLM PDG for each network is constructed from merging 18 individual LLM responses: 3 LLM models x 2 prompt detail levels x 3 repetitions using Algorithm~\ref{alg:merge_graphs}.

\subsection{Generating the BNSL Ensemble PDG and Final Fused PDG}

Three contrasting statistical BNSL algorithms capable of learning discrete networks are used in this study: PC-Stable, Tabu-Stable and FGES.\hl{ These offer different learning approaches to encourage structural diversity in the BNSL ensemble. }\change{\textbf{rev1:} emphasise diversity rationale for algorithm choice} PC-Stable \citep{colombo2014order} is a constraint-based algorithm which learns a CPDAG\footnote{Inconsistent results from statistical independence tests mean that PC-Stable sometimes returns Partially Directed Acyclic Graph (PDAG); a mixed directed and undirected graph which is not extendable to a DAG \citep{dor1992simple}}, and we use the implementation provided by bnlearn \citep{bnlearn}. Tabu-Stable \citep{kitson2025stable} is a recent greedy score-based algorithm which learns a DAG which performed well against other BNSL algorithms. FGES \citep{ramsey2017million} is a score-based algorithm which searches equivalence class space \hl{efficiently}\deleted{ and returns}\hl{, returning}\change{\textbf{rev1:} highlight efficiency of FGES} a CPDAG; we use the Tetrad implementation \citep{ramsey2018tetrad}. We primarily use a sample size of 10,000 rows, but results with 1,000 rows are included in the ablation studies in Subsection~\ref{sub:ablation}.

The upper part of Figure~\ref{fig:pdg_merge} illustrates how the BNSL ensemble PDG is generated. Since we wish to perform \emph{probabilistic} fusion, we combine sets of five individual learnt graphs to construct a PDG reflecting the uncertainties in BNSL across different data samples. Since \textsc{MergeGraphs} (Algorithm~\ref{alg:merge_graphs}) operates on the probabilities of directed and undirected states for each edge, it is able to merge any combination of DAGs, PDAGs and CPDAGs that may have been learnt. We produce PDGs based on three different sets of five datasets for each BNSL algorithm and in turn merge these individual PDGs to create one final ensemble BNSL PDG which encapsulates the diversity across dataset samples and algorithms. Finally, we use \textsc{MergeGraphs} to merge the BNSL and LLM ensemble PDGs into one fused PDG from which a causal DAG is obtained using \textsc{ToDagGreedy} (Algorithm~\ref{alg:to_dag_greedy}).

\subsection{Evaluation}

We evaluate performance on 26 small and medium-sized discrete variable networks with between 8 and 70 variables, obtained from the bnlearn \citep{bnrepository}, bnRep \citep{leonelli2025bnrep} and Bayesys \citep{bayesysrepository} network repositories. These cover a wide range of application domains and each have an associated publication justifying the network structure. We choose discrete networks as they make fewer assumptions about the variable distributions. Further details of the networks used are provided in Appendix~\ref{apdx:networks}.

As noted previously, the BN definition and associated publication are used to generate the context for the LLMs, and the reference DAG and probability tables defined by the former are used to generate clean synthetic observational data samples that the BNSL algorithms learn structures from. Output DAGs are \hl{primarily} evaluated against the reference DAG using the F1 metric since our focus here is on causal discovery; we follow the semantics for computing F1 that is used in bnlearn \citep{bnlearn}.\hl{ However, some results are reported using the Structural Hamming Distance (SHD) metric on both DAGs and CPDAGs for comparison with other studies.}\change{\textbf{rev3: }some results using SHD included.}

\section{Results}
\label{sec:results}

\subsection{Analysing Edge Diversity in the LLM and BNSL PDGs}

\begin{figure}[!ht]
\centering
\includegraphics[width=1.0\textwidth]{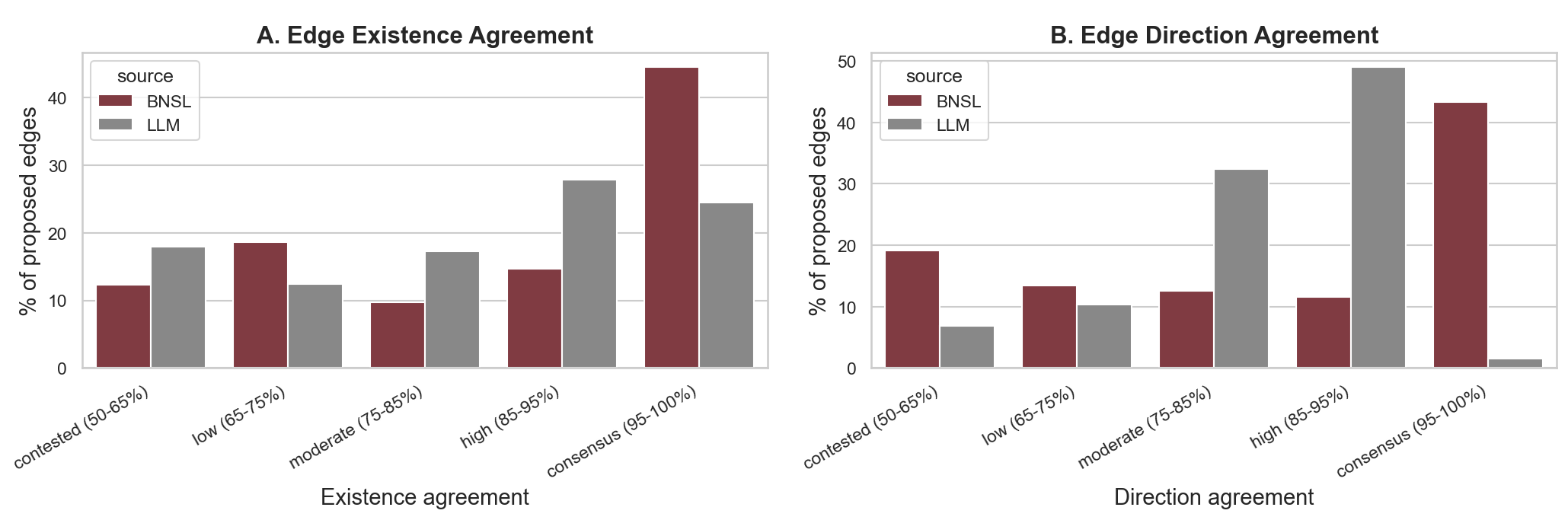}
\caption{Percentage of edges in agreement as regards existence and orientation for LLM and BNSL sources. The x-axis categorises the probability of the more likely outcome for each edge: whether the edge exists or not, and its direction. The y-axis shows the percentage of edges falling in each category.}
\label{fig:edge_diversity}
\end{figure}

We first analyse the diversity in edge orientation and existence in our inputs to the fusion process because, as Subsection~\ref{sub:ablation} shows, this variability contributes to the success of our approach. The grey bars in Figure~\ref{fig:edge_diversity} show the diversity in edge existence and edge orientation across the individual LLM PDGs. The horizontal axis categorises the mean probability of the majority outcome for an edge. For example, if the mean value for $p_{none} = 0.4$ for a particular edge, the majority outcome is that the edge exists ($p = 0.6$), and its existence would be categorised as "contested". We see that there is a considerable proportion of edges where LLM responses disagree about existence or orientation, and in particular, fewer edges where there is consensus on edge existence than BNSL provides.

The maroon bars in Figure~\ref{fig:edge_diversity} illustrate the edge diversity produced across the individual BNSL PDGs. These have more consensus on edge existence than LLMs, but less for edge orientation with nearly 20\% of edges having contested orientation. However, BNSL produces many more consensus orientations than LLMs and this may be partly due to colliders being identified by all algorithms; nearly 50\% of collider edges reached consensus compared to 35\% of non-collider edges.

\subsection{Improvement in Structural Accuracy}

\begin{table}[!ht]
  \centering
  \caption{Summary of fusion improvement in F1 over
    BNSL and LLM alone (26 networks, 10K samples).
    Wilcoxon signed-rank two-sided test.}
  \label{tab:fusion-summary}
  \begin{tabular}{lccccc}
    \toprule
    Comparison & Wins & Losses
      & Mean $\Delta$F1 & $W$ & $p$ \\
    \midrule
    Fusion vs.\ BNSL
      & 23 & 3 & +0.099 & 17.0 & $<$0.001 \\
    Fusion vs.\ LLM
      & 25 & 1 & +0.188 & 3.0 & $<$0.001 \\
    Fusion vs.\ best of either
      & 22 & 4 & +0.056 & 36.0 & $<$0.001 \\
    \bottomrule
  \end{tabular}
\end{table}

Table~\ref{tab:fusion-summary} provides a summary of the gains provided by BNSL/LLM fusion compared to using either BNSL or LLM alone. The BNSL and LLM metrics are produced by converting the ensemble BNSL and LLM PDGs into DAGs using Algorithm~\ref{alg:to_dag_greedy} and comparing these with the reference DAG. The ensemble LLM and BNSL PDG are combined using Algorithm~\ref{alg:merge_graphs} to create a fused PDG which is then similarly converted to a DAG.

The fused DAG has a higher F1 than the BNSL DAG in 23/26 networks, with a mean F1 improvement of 0.099 which is significant at
$p<0.001$ (Wilcoxon signed-rank two-sided test, $W=17.0$). The fused DAG achieves a higher F1 score than the LLM DAG in 25/26 networks with a mean improvement of 0.188, also significant at $p < 0.001$, $W=3.0$. When the fused DAG F1 is compared to the better of LLM and BNSL, it is better in 22/26 networks with a significant mean improvement of 0.056 ($p<0.001$, $W=36.0$). A breakdown of these results by network is provided in Table~\ref{tab:fusion} in Appendix~\ref{apdx:detailed_results}. \hl{Direct empirical comparison with prior LLM-assisted causal discovery methods is difficult because studies use different benchmark networks, sample sizes, and evaluation protocols.}\change{\textbf{rev2:} acknowledge difficulties comparing results with other studies.}

\begin{figure}[htbp]
\centering
\includegraphics[width=0.9\textwidth]{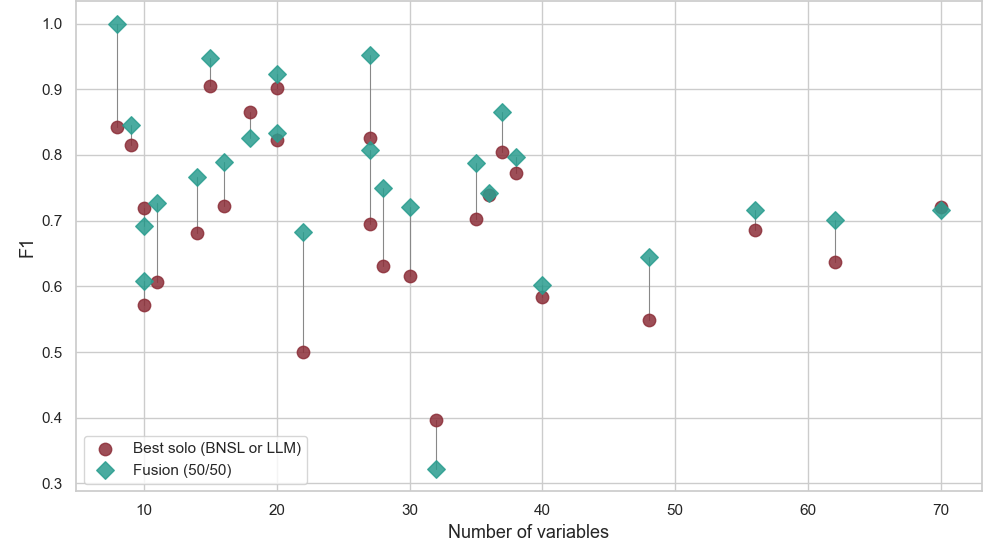}
\caption{Change in F1 moving from the best of BNSL and LLM to the fused DAG plotted against number of variables in the network.}
\label{fig:fusion_vs_solo}
\end{figure}

Figure~\ref{fig:fusion_vs_solo} plots the change in F1 moving from the best of LLM and BNSL to the fused DAG. Fusion delivers improvements across a range of network sizes, with the most consistent gains for medium-sized networks (15 - 40 variables). For very small networks, individual improvements can be large but are less consistent, while for the largest networks the gains are more modest, possibly reflecting the limits of LLM causal reasoning over large variable sets.

\begin{figure}[htbp]
\centering
\includegraphics[width=1.0\textwidth]{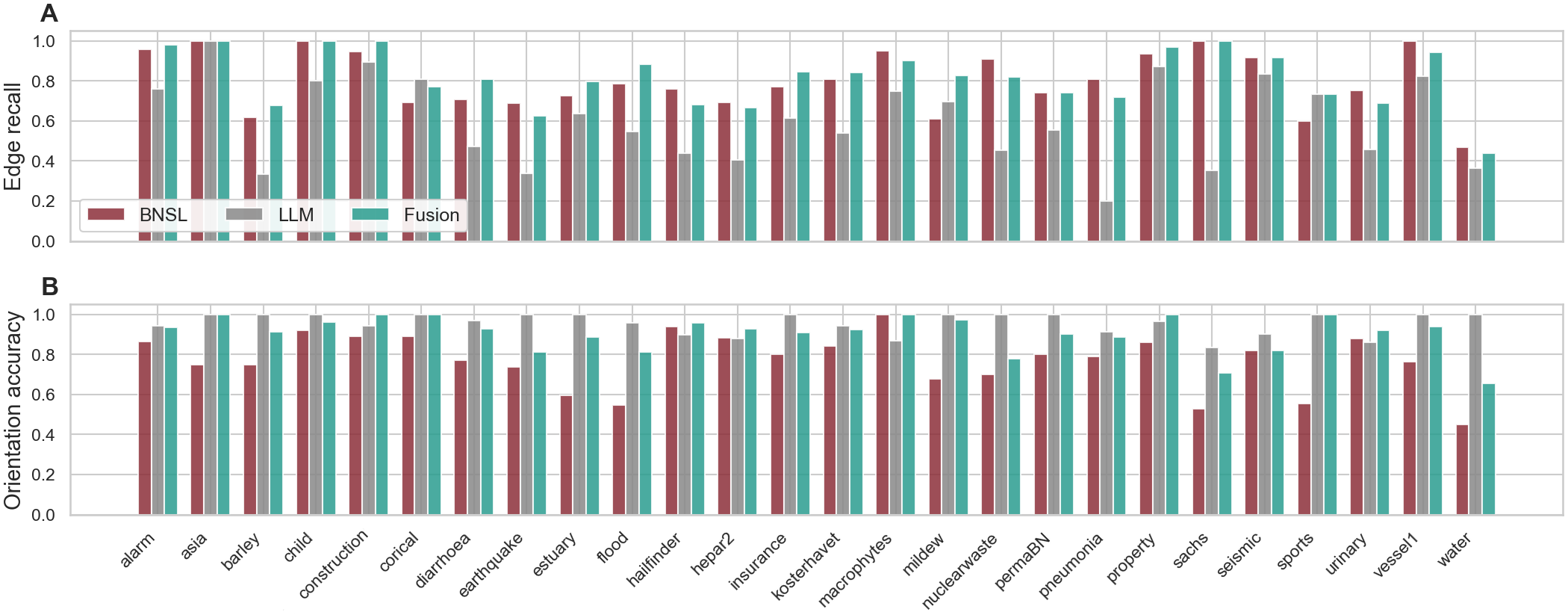}
\caption{\textbf{(A)} Edge-recall (fraction of reference arcs detected at $p_{exist} \geq 0.3$) and \textbf{(B)} orientation accuracy (fraction of detected reference arcs correctly directed) for BNSL, LLM, and the fused PDG across 26 benchmark networks.}
\label{fig:recall_orientation}
\end{figure}

To understand \emph{why} fusion improves over either source alone, we decompose the contribution of each source into edge existence and edge orientation. For each true arc in the reference graph, we ask: (i) does the source detect it, and (ii) does it assign the correct direction? BNSL has substantially higher edge-recall than LLM at 0.802 compared to 0.603 for LLM (Figure~\ref{fig:recall_orientation}(A)). While most reference arcs are detected by both sources, many more are found exclusively by BNSL than exclusively by LLM: 445 versus 103 across all 26 networks, a ratio of more than four to one. Fusion retains 379/445 of these BNSL-only arcs, achieving a mean recall of 0.818, slightly above BNSL alone. BNSL's higher recall thus contributes more to the fused skeleton than LLM, particularly for edges that only one source detects.

While BNSL leads at finding edges, LLM leads at orientating them, a finding consistent with \cite{constantinou2025using} and \cite{wan2025large}, for example. Figure~\ref{fig:recall_orientation}(B) reports orientation accuracy, the fraction of detected reference arcs assigned the correct direction. LLM achieves a mean orientation accuracy of 0.96, substantially higher than BNSL's 0.77. This gap is expected: constraint-based algorithms such as PC, and equivalence search algorithms such as FGES produce partially directed graphs with many undirected edges, whereas LLMs express confident directional beliefs. Among the 312 reference arcs where BNSL either favoured the wrong direction or had weak orientation clarity, fusion corrected 210 (67\%). Moreover, the per-network gain in orientation accuracy from BNSL to fusion correlates significantly with the overall F1 improvement (Spearman $r=0.51$, $p=0.008$), confirming that orientation correction is a primary mechanism through which LLM knowledge improves the fused graph.

Neither edge existence complementarity nor probability-weighted divergence between sources showed a significant relationship with per-network F1 improvement (Spearman $r=0.28$, $p=0.17$ and $r=0.23$, $p=0.25$ respectively), nor did any of the existence metrics alone (all $p>0.07$). In contrast, the orientation accuracy gain is a significant predictor ($r=0.51, p=0.008$). Taken together, these results suggest that the two sources play complementary but distinct roles: BNSL contributes a high-recall edge skeleton from data, while LLM contributes accurate orientation from domain knowledge. Neither contribution alone explains fusion's benefit. It is the combination of data-driven edge detection with knowledge-driven edge orientation that yields the consistent F1 improvements seen.

\hl{To investigate}\change{\textbf{pc3:} Added this analysis of what proportion of false positive edges in LLM and BNSL graphs correspond to reversed edges, mediated edges, common cause edges as well as truly spurious ones.}\hl{ whether LLMs distinguish direct causal relationships from mediated or associational relationships, we categorised LLM false-positive edges. Only 7\% corresponded to reversed true causal edges, whereas 31\% were mediated effects and 30\% involved variables sharing a common cause. This suggests that LLMs often recognise causally related variables but do not always distinguish direct causation from indirect or common-cause relationships. In contrast, the dominant BNSL error mode was incorrect orientation (60\% of false positives), further highlighting the complementary nature of the two sources.}

\hl{We additionally evaluate structural accuracy using SHD. For DAG comparisons, fusion reduced mean SHD from 24.7 (BNSL) and 43.4 (LLM) to 22.3, being superior to the better of the two sources in 16 of 26 networks. The improvement is less pronounced than for F1, which is consistent with our finding that fusion primarily contributes through improved edge orientation which affects F1 more. Full results are provided in Appendix~}\ref{apdx:detailed_results}\hl{. We additionally evaluated CPDAG SHD and observed smaller gains. This is expected because CPDAG representations remove many orientation differences, reducing structural distances overall. Importantly, the benefits of fusion remain evident at the equivalence-class level.}\change{\textbf{rev3:} discuss SHD results}

The BNSL baseline used throughout this study is itself an ensemble of three complementary algorithms (FGES, Tabu, PC). This BNSL ensemble \emph{already improves} F1 by a mean of +0.062 over the average\change{have removed table that showed exactly these results to save space} individual algorithm ($p=0.003$, Wilcoxon signed-rank test), with gains in 20 of 26 networks (per-network detail in Table~\ref{tab:bnsl-ensemble-detail} in Appendix~\ref{apdx:bnsl-ensemble-detail}). The fusion gains reported above are therefore measured against this \emph{strengthened baseline}. Were fusion instead compared to individual BNSL algorithms, the mean improvement would rise from +0.099 to +0.161 ($p<0.001$). This underscores that fusion with LLM knowledge provides substantial benefit beyond that which conventional graph averaging alone can achieve.

\subsection{Ablation Studies}
\label{sub:ablation}

\begin{table}[ht]
\centering
\begin{tabular}{lccccc}
\toprule
Ablation Type & Ablation 1 & Ablation 2 & Ablation 3 & Diversity Gain \\
\midrule
LLM Model      & Gemini: 0.700 & Claude: 0.683 & \textbf{GPT: 0.714}  & +0.046 \\
BNSL Algorithm & \textbf{FGES: 0.721}   & Tabu: 0.672   & PC: 0.654  & +0.039 \\
LLM Repetition & First: 0.727  & \textbf{Second: 0.734}  & Third: 0.725  & +0.026 \\
Prompt Detail  & Minimal: 0.696 & \textbf{Standard: 0.730} & ---      & +0.030 \\
Sample Size    & 1K: 0.674      & \textbf{1K+10K: 0.729} & ---       & \textbf{-0.031} \\
\bottomrule
\end{tabular}
\caption{Ablation results (F1 score) for each diversity axis. All values use 10K data except for the sample size row, where 1K and 1K+10K are shown. For each axis, the best single ablation is bolded. Diversity gain is the difference between the main result (10K-only, F1=0.760) and the best ablation in each row.}
\label{tab:ablation-summary}
\end{table}

Table~\ref{tab:ablation-summary} provides a comparison of ablation results across several axes. For each axis, we report the F1 score for each ablation case, where we restrict the fused result to just using one of the ablation values, e.g., in the first row, only using Gemini rather than all LLM models. The final column shows the diversity gain; that is, the improvement in F1 that comes from using all variants compared to just using the best performing individual variant.

For prompt detail, the standard prompt (\textbf{0.730}) outperforms the minimal prompt (0.696), indicating that more detailed prompts yield better results. Among LLMs, GPT (\textbf{0.714}) is the best individual model, slightly ahead of Gemini (0.700) and Claude (0.683). For BNSL algorithms, FGES (\textbf{0.721}) is superior to Tabu (0.672) and PC (0.654). For LLM repetition, the three identical requests unsurprisingly obtain similar levels of accuracy. However, in all these cases, the ensemble using all variants further improves performance over the best individual variant, as shown by the positive diversity gain.

The exception is sample size: while the ensemble of 1K and 10K (0.729) is better than 1K alone (0.674), it is outperformed by using only the largest sample (10K, F1=0.760), resulting in a negative diversity gain. This suggests that, while diversity due to using different LLM models and prompt levels, repetitive high-temperature requests to LLMs, and using different BNSL algorithms all improve performance, maximising data size is more important than sample size diversity here.

Our method weights the BNSL and LLM sources equally, but Table~\ref{tab:oracle} in Appendix~\ref{apdx:detailed_results} explores whether an optimal per-network weighting could improve results further. We assign weights $(b,1-b)$ to BNSL and LLM respectively, vary $b \in [0, 1]$ in steps of $0.05$, and select the "Oracle" weighting that maximises F1 against the reference graph. Table~\ref{tab:oracle} reports the best BNSL weight and the resulting F1 gain over 50/50 fusion for each network. Oracle weighting improves F1 by a mean of 0.027 over equal weighting, and by 0.083 over the better of BNSL and LLM alone, with 25/26 networks now exceeding both solo sources. While impractical without a known reference graph, this suggests that an objective method of determining $b$ would yield further gains\footnote{Early experiments using edge variability or BIC score of the final DAG to select the weighting were not successful.}. However, in the absence of a principled method of determining $b$, using a value of $0.5$ represents a reasonable and robust \hl{heuristic approach}.\change{\textbf{rev5:} emphasise that 50/50 is a heuristic approach}

\subsection{Post-hoc Information Leakage Analysis}
\label{sub:leakage}

\hl{To }\change{\textbf{pc1:} new section added investigating benchmark identification and causal information leakage from variable names etc.}\hl{investigate benchmark recognition and answer-aware context generation, we performed a post-hoc audit of all 26 network context files used to generate prompts for the LLMs. For each network, Claude Opus 4.5 assessed (i) benchmark identifiability and (ii) the extent to which variable names and descriptions suggested causal structure. Further details and per-network results are provided in Appendix~}\ref{apdx:leakage}\hl{.}

\hl{The variable name obfuscation substantially reduced benchmark identifiability for several networks. Moreover, neither benchmark identifiability nor estimated causal information leakage was associated with the improvement obtained from fusion. Benchmark-identification scores had only weak correlations with fusion improvement (Spearman $\rho = 0.14$ for minimal and $\rho = 0.18$ for standard prompts), whilst causal-leakage scores were similarly unrelated ($\rho = -0.19$ and $\rho = -0.09$ respectively; all $p > 0.35$). Estimated leakage scores changed little between minimal and standard prompts and the corresponding performance difference was modest (F1 gain = 0.696 vs. 0.730), suggesting that variable descriptions, types and states contribute limited additional information.}

\hl{Qualitative inspection suggested that most inferred causal information arose from the ordinary semantics of the variables themselves (e.g., rainfall, runoff, photosynthesis, biomass and yield) rather than benchmark-specific identifiers or explicit structural descriptions. While benchmark recognition cannot be ruled out, these analyses provide no evidence that benchmark recognisability or causal information leakage explains the observed variation in fusion gains across networks.}

\section{Conclusion}
\label{sec:conclusion}

We have presented a straightforward approach to combining data-driven BNSL with LLM-derived causal knowledge through PDGs. By representing both sources as probability distributions over edge states and merging them using weighted averaging, we obtain fused graphs that consistently outperform either source in isolation. Across 26 benchmark networks, equal-weight fusion improves F1 over the better of BNSL and LLM alone in 22 of 26 cases, with a mean improvement in F1 of 0.056 that is statistically significant ($p<0.001$, Wilcoxon signed-rank test). The approach requires no modifications to the underlying structure learning algorithms and uses standard LLM APIs and models.

Our analysis reveals that the two sources play complementary roles. BNSL provides a high-recall edge skeleton, detecting 80\% of
reference arcs compared to 60\% for LLM, while LLMs provide accurate edge orientation, achieving 96\% orientation accuracy
compared to 77\% for BNSL. The per-network gain in orientation accuracy is a significant predictor of F1 improvement ($r=0.51$,
$p=0.008$), confirming that orientation correction is the primary mechanism through which LLM knowledge benefits the fused graph.
Ablation studies show that diversity across LLM models, prompt detail levels, and repeated high temperature requests, as well as across BNSL algorithms, each contribute positively to improved accuracy, but maximising data sample size is more important than sample size diversity.

The current study has several limitations. We evaluate only on discrete-variable networks of up to 70 nodes; applicability to
larger or continuous-variable networks remains to be tested. \hl{We do not assess probability calibration, and therefore PDG values should not presently be interpreted as true posterior probabilities. }\change{\textbf{rev5:} acknowledge that PDG probabilities cannot be treated as true posteriors.} The equal-weight fusion, while effective, leaves room for improvement:
oracle analysis shows a mean additional gain of 0.027 from per-network optimal weighting.\hl{ We also evaluated the potential impact of benchmark recognition and information leakage through a post-hoc audit of all network contexts. Whilst neither benchmark identifiability nor estimated causal information leakage was associated with fusion improvement, these remain important threats to validity that warrant further investigation.}\change{\textbf{pc2:} report that benchmark identification and causal leakage were investigated, but remain significant threats.} Finally, we use clean synthetic data \deleted{without measurement noise or latent confounders}\hl{ from fully observed networks}. Whilst this allows controlled evaluation against known ground truth DAGs\hl{, it does not assess robustness to measurement noise, missing data, latent confounding, or model misspecification, all of which are common in real-world applications and so our study}\change{\textbf{rev6:} emphasise the limitations of using clean synthetic data} likely overestimates the performance of BNSL in real-world applications. In such settings, the complementary LLM-based knowledge could play an even greater role in improving accuracy.

Several directions for future work arise from these findings. First, the PDG framework can accommodate additional
knowledge sources including expert elicitation, and transfer from related domains. Improvement to both the LLM and BNSL components through a wider range of models and algorithms could be explored, particularly state-of-the-art offerings. Development of an objective BNSL/LLM weighting criterion could improve results. Extending the evaluation to continuous and mixed-variable networks, as well as larger networks would be valuable, as would evaluating robustness to data noise and latent variables. The latter would be of particular interest as fusion might be particularly valuable in that case. \hl{The current PDG representation assumes causal sufficiency and lacks an explicit bidirected edge state to represent latent confounding. Consequently, confounded relationships cannot be represented directly and may instead appear as common-cause or incorrectly directed dependencies. Extending PDGs with a bidirected state is an important direction for future work.}\change{\textbf{pc3:} add a discussion of PDGs supporting bidirected edges}\hl{ Further work to mitigate benchmark identification, including the development of new theoretically grounded benchmarks, would strengthen future evaluations.}\change{\textbf{pc2}: note further work needed to prevent benchmark identification is needed.} Finally, more sophisticated fusion strategies, such as querying LLMs only for edges where BNSL is uncertain, could improve efficiency and further exploit the complementary strengths identified in this study.


\newpage
\bibliography{references}
\newpage

\reversemarginpar
\setlength{\marginparwidth}{2cm} 

\appendix

\section{LLM Prompts and Responses}
\label{apdx:llm_prompt}

Figure~\ref{fig:llm_prompt_example} shows an example request and response sent to LLMs to obtain edge and orientation probabilities for the Asia benchmark network. The input request contains a system prompt defining the LLM's role, and a user prompt specifying the question directed at the LLM. Note that the standard benchmark name, Asia, is not mentioned in the request, and variables are renamed from those in the standard benchmark network definition to try and avoid the LLM regurgitating the standard network definition.

\begin{figure}[H]
\centering
\fbox{\parbox{0.92\columnwidth}{%
\ttfamily\footnotesize
{\sffamily\bfseries SYSTEM PROMPT:}\\
You are an expert in causal reasoning. Propose causal relationships with uncertainty estimates.

Respond with ONLY this JSON structure:\\
\hspace*{1em}{"edges":[{"source":"X","target":"Y", "existence":0.9,"orientation":0.8}]}

Rules:\\
- source: cause variable name (exact spelling) \\
- target: effect variable name (exact spelling) \\
- existence: probability a relationship exists \\
- orientation: confidence source causes target \\
- ONLY include edges with existence >= 0.5 \\
- ONLY direct causal relationships \\

{\sffamily\bfseries USER PROMPT:}\\
In the domain of respiratory\_diagnosis:\\
\\
Propose a causal graph for the following variables:\\
\\
- endemic\_travel\\
\hspace*{1em}Type: binary\\
\hspace*{1em}Description: Whether patient has recently travelled to a tuberculosis-endemic region.\\
\hspace*{1em}States: no, yes\\
\\
- mycobacterial\_infection\\
\hspace*{1em}Type: binary\\
\hspace*{1em}Description: Whether patient has active tuberculosis infection.\\
\hspace*{1em}States: no, yes\\
\\
\hspace*{2em}...\\
\\
{\sffamily\bfseries RESPONSE:}\\
\{"edges":[\\
\hspace*{1em}\{"source":"endemic\_travel",\\
\hspace*{1.5em}"target":"mycobacterial\_infection",\\
\hspace*{1.5em}"existence":0.95,\\
\hspace*{1.5em}"orientation":1.0\\
\hspace*{1em}\},
\\
\hspace*{2em}...\\
\\
\hspace*{1em}]\\
\}
}}
\caption{Example LLM prompt and response for the
  Asia network (standard detail level).}
\label{fig:llm_prompt_example}
\end{figure}

\begin{table}[!htbp]
  \centering
  \caption{LLM request and response sizes, costs and latency over the 468 LLM requests in this study.}
  \label{tab:llm_requests}
  \begin{tabular}{lllllllll}
    \toprule
     &  \multicolumn{2}{c}{Request Tokens} & \multicolumn{2}{c}{Response Tokens} & \multicolumn{2}{c}{Cost (US cents)} & \multicolumn{2}{c}{Latency (secs)} \\
LLM    &  Mean &  Max. &  Mean &  Max. &  Mean &  Max. &  Mean &  Max.  \\
\midrule
claude &      964 &    3327 &     1680 &    5239 &    0.00 &   0.00 &     8.0 &   22.3 \\
gemini    &      898 &    3136 &     2464 &    9132 &    0.00 &   0.00 &    37.9 &  122.9 \\
gpt    &      850 &    3049 &      793 &    1913 &    0.16 &   0.41 &    12.5 &   55.3 \\
\midrule
All       &      904 &    3327 &     1646 &    9132 &    0.05 &   0.41 &    19.4 &  122.9 \\
    \bottomrule
  \end{tabular}
\end{table}

Table~\ref{tab:llm_requests} summarises the 156 requests made to each of the three LLM models in this study. The size of requests and responses are reported by the number of tokens used which the vendor API reports. This will generally rise as the number of variables in the network increases. These LLMs enforce a limit on the number of tokens in the response after which the response is truncated: 32,768 for gpt, 16,384 for claude and 65,536 for gemini. The cost is also reported by the vendor API and shown here in U.S. cents; only gpt of the models used was not available on a free-tier at the time of the experiments (April 2026). Latency is in seconds as recorded by the programme making the API request. Once again, larger networks tend to have larger latencies.

\section{Networks used in the evaluation}
\label{apdx:networks}

Table~\ref{tab:networks} shows the 26 discrete variable networks used in this study. They were selected on the basis of having a defined reference BN on a publicly accessible repository, where the structure had been defined by expert elicitation or a literature review. The three repositories used are bnlearn \citep{bnrepository}, Bayesys \citep{bayesysrepository} and bnRep \citep{leonelli2025bnrep}, the third column showing which repository was used to obtain each network.

The first column in the table is the canonical name of the network used on the repository and is usually how the network is referred to in the literature; but as noted previously, this name is not presented to the LLM. The graph structure in the published definitions is used as the reference DAG, though variable name aliases are used in the LLM requests. The benchmark BNs also define the Conditional Probability Tables (CPTs) for each variable which together with the structure are used to generate the synthetic data used as input to the BNSL algorithms. The reference CPTs for some networks were adjusted to prevent degenerate single-valued variables occurring in the synthetic datasets. This was done for 12 CPTs in corical, 2 in barley and 6 in water. This was considered acceptable given that the focus of this study is identifying good causal \emph{structures}.

\begin{table}[H]
  \centering
  \caption{Benchmark networks used in the evaluation.}
  \label{tab:networks}
  \begin{tabular}{p{0.20\textwidth} p{0.40\textwidth} p{0.09\textwidth} r r}
    \toprule
    Network & Domain & Source & Variables & Arcs\\
    \midrule
    alarm & patient monitoring & bnlearn & 37 & 46 \\
    asia & respiratory diagnosis & bnlearn & 8 & 8 \\
    barley & crop yield prediction & bnlearn & 48 & 84 \\
    child & neonatal cardiology & bnlearn & 20 & 25 \\
    constructionprod. & construction management & bnRep & 18 & 19 \\
    corical & vaccine risk benefit analysis & bnRep & 20 & 26 \\
    diarrhoea & childhood health epidemiology & Bayesys & 28 & 68 \\
    earthquake & maritime transportation risk & bnRep & 40 & 77 \\
    estuary & estuary ecological health & bnRep & 30 & 44 \\
    flood & natural flood management \newline agriculture & bnRep & 22 & 42 \\
    hailfinder & severe weather forecasting & bnlearn & 56 & 66 \\
    hepar2 & hepatological diagnosis & bnlearn & 70 & 123 \\
    insurance & automobile insurance risk & bnlearn & 27 & 52 \\
    kosterhavet & marine ecosystem management & bnRep & 38 & 63 \\
    macrophytes & freshwater ecosystem macrophyte \newline management & bnRep & 15 & 20 \\
    mildew & crop disease modelling & bnlearn & 35 & 46 \\
    nuclearwaste & nuclear waste repository safety & bnRep & 10 & 11 \\
    permaBN & arctic permafrost thaw prediction & bnRep & 14 & 27 \\
    pneumonia & paediatric respiratory infection & bnRep & 62 & 171 \\
    property & property investment analysis & bnlearn & 27 & 31 \\
    sachs & protein signalling & bnlearn & 11 & 17 \\
    seismic & structural seismic risk & bnRep & 10 & 12 \\
    sports & football match prediction & Bayesys & 9 & 15 \\
    urinary & paediatric urology & bnRep & 36 & 109 \\
    vessel1 & fishing vessel sinking accidents & bnRep & 16 & 17 \\
    water & water treatment monitoring & bnlearn & 32 & 66 \\
    \bottomrule
  \end{tabular}
\end{table}

\section{Detailed per-network Fusion Results}
\label{apdx:detailed_results}

The BNSL and LLM columns in Table~\ref{tab:fusion} report the F1 of the DAG derived from merging all PDGs within each source as described in Subsection~\ref{sub:pdg}. The Fusion column reports the F1 from a two-stage merge: first, PDGs within each source are merged into a single BNSL and a single LLM consensus PDG; these are then combined with equal (50/50) weighting. This design ensures each source contributes equally to the fused result, regardless of the number of constituent PDGs. A DAG is derived from this fused PDG and its F1 reported in the penultimate column of the Table. The final column of Table~\ref{tab:fusion} reports the improvement in F1 offered by the fused DAG compared to the better of BNSL and LLM alone.

\begin{table}[H]
  \centering
  \caption{Improvement in F1 attained by fusing BNSL and LLM over the better of BNSL and LLM alone.}
  \label{tab:fusion}
  \begin{tabular}{p{0.26\textwidth} p{0.1\textwidth} lll p{0.18\textwidth}}
    \toprule
    Network & Nodes & BNSL & LLM & Fusion & Fusion improvement over the better of BNSL and LLM \\
    \midrule
alarm                     &    37 &   0.804 &   0.653 &   0.866 & +0.062 \\
asia                      &     8 &   0.750 &   0.842 &   1.000 & +0.158 \\
barley                    &    48 &   0.549 &   0.392 &   0.645 & +0.096 \\
child                     &    20 &   0.902 &   0.702 &   0.923 & +0.021 \\
constructionproductivity  &    18 &   0.865 &   0.525 &   0.826 & -0.039 \\
corical                   &    20 &   0.727 &   0.824 &   0.833 & +0.010 \\
diarrhoea                 &    28 &   0.632 &   0.456 &   0.750 & +0.118 \\
earthquake                &    40 &   0.584 &   0.382 &   0.602 & +0.018 \\
estuary                   &    30 &   0.435 &   0.615 &   0.721 & +0.106 \\
flood                     &    22 &   0.474 &   0.500 &   0.682 & +0.182 \\
hailfinder                &    56 &   0.686 &   0.409 &   0.717 & +0.031 \\
hepar2                    &    70 &   0.721 &   0.421 &   0.717 & -0.004 \\
insurance                 &    27 &   0.695 &   0.674 &   0.808 & +0.113 \\
kosterhavet               &    38 &   0.772 &   0.538 &   0.797 & +0.025 \\
macrophytes               &    15 &   0.905 &   0.619 &   0.947 & +0.043 \\
mildew                    &    35 &   0.519 &   0.703 &   0.787 & +0.084 \\
nuclearwaste              &    10 &   0.571 &   0.417 &   0.609 & +0.037 \\
permaBN                   &    14 &   0.681 &   0.625 &   0.766 & +0.085 \\
pneumonia                 &    62 &   0.637 &   0.253 &   0.702 & +0.065 \\
property                  &    27 &   0.825 &   0.812 &   0.952 & +0.127 \\
sachs                     &    11 &   0.606 &   0.400 &   0.727 & +0.121 \\
seismic                   &    10 &   0.720 &   0.692 &   0.692 & -0.028 \\
sports                    &     9 &   0.400 &   0.815 &   0.846 & +0.031 \\
urinary                   &    36 &   0.738 &   0.522 &   0.742 & +0.003 \\
vessel1                   &    16 &   0.722 &   0.683 &   0.789 & +0.067 \\
water                     &    32 &   0.283 &   0.397 &   0.322 & -0.075 \\
\midrule
AVERAGE                   &       &   0.662 &   0.572 &   0.760 & +0.056 \\
    \bottomrule
  \end{tabular}
\end{table}

\hl{Table~}\ref{tab:fusion-shd}\hl{ shows the gains in DAG accuracy resulting from fusion according to the SHD metric. The mean improvement (decrease) in SHD was $-1.7$ with SHD improving (decreasing) in 15 / 26 networks. The improvement is less pronounced than indicated by F1, which is consistent with our finding that the LLM contributes primarily through better orientation; the F1 metric penalises misorientation more strongly than SHD.}\change{\textbf{rev3: }DAG SHD comparison added.}

\begin{table}[H]
  \centering
  \caption{Change in DAG SHD attained by fusing BNSL and LLM over the better of BNSL and LLM alone.}
  \label{tab:fusion-shd}
  \begin{tabular}{p{0.26\textwidth} p{0.1\textwidth} lll p{0.18\textwidth}}
    \toprule
    Network & Nodes & BNSL & LLM & Fusion & Fusion DAG \newline SHD change over the better of BNSL and LLM \\
    \midrule
alarm                     &    37 &  12.000 &  33.000 &  11.000 & -1.000  \\
asia                      &     8 &   2.000 &   3.000 &   0.000 & -2.000  \\
barley                    &    48 &  59.000 &  87.000 &  50.000 & -9.000  \\
child                     &    20 &   3.000 &  17.000 &   3.000 &  0.000  \\
constructionproductivity  &    18 &   3.000 &  28.000 &   8.000 & +5.000  \\
corical                   &    20 &  10.000 &   9.000 &   8.000 & -1.000  \\
diarrhoea                 &    28 &  33.000 &  73.000 &  31.000 & -2.000  \\
earthquake                &    40 &  44.000 &  84.000 &  47.000 & +3.000  \\
estuary                   &    30 &  40.000 &  35.000 &  20.000 & -15.000 \\
flood                     &    22 &  26.000 &  43.000 &  20.000 & -6.000  \\
hailfinder                &    56 &  40.000 &  72.000 &  32.000 & -8.000  \\
hepar2                    &    70 &  48.000 & 115.000 &  55.000 & +7.000  \\
insurance                 &    27 &  22.000 &  31.000 &  15.000 & -7.000  \\
kosterhavet               &    38 &  19.000 &  53.000 &  21.000 & +2.000  \\
macrophytes               &    15 &   4.000 &  14.000 &   2.000 & -2.000  \\
mildew                    &    35 &  32.000 &  27.000 &  19.000 & -8.000  \\
nuclearwaste              &    10 &   6.000 &  14.000 &   8.000 & +2.000  \\
permaBN                   &    14 &  11.000 &  18.000 &   9.000 & -2.000  \\
pneumonia                 &    62 &  95.000 & 180.000 &  80.000 & -15.000 \\
property                  &    27 &   8.000 &  11.000 &   3.000 & -5.000  \\
sachs                     &    11 &   7.000 &  14.000 &   7.000 & 0.000  \\
seismic                   &    10 &   5.000 &   7.000 &   6.000 & +1.000  \\
sports                    &     9 &  11.000 &   5.000 &   4.000 & -1.000  \\
urinary                   &    36 &  41.000 &  70.000 &  43.000 & +2.000  \\
vessel1                   &    16 &   6.000 &  13.000 &   8.000 & +2.000  \\
water                     &    32 &  54.000 &  73.000 &  70.000 & +16.000 \\
\midrule
AVERAGE                   &       &  24.654 &  43.423 &  22.308 & -1.692 \\
    \bottomrule
  \end{tabular}
\end{table}

\begin{table}[H]
  \centering
  \caption{Change in CPDAG SHD attained by fusing BNSL and LLM over the better of BNSL and LLM alone.}
  \label{tab:fusion-cpdag}
  \begin{tabular}{p{0.26\textwidth} p{0.1\textwidth} lll p{0.18\textwidth}}
    \toprule
    Network & Nodes & BNSL & LLM & Fusion & Fusion CPDAG \newline SHD change over the better of BNSL and LLM \\
    \midrule
alarm                     &    37 &  13.000 &  38.000 &  11.000 & -2.000 \\
asia                      &     8 &   0.000 &   4.000 &   0.000 & 0.000 \\
barley                    &    48 &  69.000 &  91.000 &  59.000 & -10.000 \\
child                     &    20 &   1.000 &  22.000 &   5.000 & +4.000 \\
constructionproductivity  &    18 &   1.000 &  33.000 &  11.000 & +10.000 \\
corical                   &    20 &  13.000 &  15.000 &  15.000 & +2.000 \\
diarrhoea                 &    28 &  47.000 &  79.000 &  51.000 & +4.000 \\
earthquake                &    40 &  48.000 &  91.000 &  56.000 & +8.000 \\
estuary                   &    30 &  42.000 &  39.000 &  23.000 & -16.000 \\
flood                     &    22 &  29.000 &  48.000 &  23.000 & -6.000 \\
hailfinder                &    56 &  48.000 &  73.000 &  34.000 & -14.000 \\
hepar2                    &    70 &  48.000 & 119.000 &  56.000 & +8.000 \\
insurance                 &    27 &  36.000 &  34.000 &  26.000 & -8.000 \\
kosterhavet               &    38 &  19.000 &  56.000 &  22.000 & +3.000 \\
macrophytes               &    15 &   4.000 &  15.000 &   4.000 & 0.000 \\
mildew                    &    35 &  33.000 &  27.000 &  19.000 & -8.000 \\
nuclearwaste              &    10 &   5.000 &  14.000 &   8.000 & +3.000 \\
permaBN                   &    14 &  10.000 &  20.000 &   9.000 & -1.000 \\
pneumonia                 &    62 & 101.000 & 181.000 &  86.000 & -15.000 \\
property                  &    27 &   8.000 &  16.000 &   4.000 & -4.000 \\
sachs                     &    11 &   1.000 &  17.000 &  17.000 & +16.000 \\
seismic                   &    10 &   4.000 &   9.000 &   6.000 & +2.000 \\
sports                    &     9 &   8.000 &   5.000 &   4.000 & -1.000 \\
urinary                   &    36 &  49.000 &  73.000 &  49.000 & 0.000 \\
vessel1                   &    16 &   2.000 &  14.000 &   9.000 & +7.000 \\
water                     &    32 &  53.000 &  76.000 &  73.000 & +20.000 \\
\midrule
AVERAGE                   &       &  26.615 &  46.500 &  26.154 & +0.077 \\
\bottomrule
  \end{tabular}
\end{table}

\hl{Table~}\ref{tab:fusion-cpdag}\hl{ reports SHD after converting graphs to CPDAGs. Overall, fusion slightly worsens CPDAG SHD when compared to the better of LLM or BNSL, with CPDAG SHD being better than both BNSL and LLM in 11 / 26 networks. This is generally because the BNSL SHD is much better than the LLM one which is expected since LLM outputs represent directed causal beliefs, whereas CPDAG metrics reward preservation of equivalence-class uncertainty. Consequently, this metric naturally favours structure-learning algorithms such as PC and FGES that operate directly in equivalence-class space.}\change{\textbf{rev3: }CPDAG SHD comparison added.}

\begin{table}[!ht]
  \centering
  \caption{Improvement in F1 using an optimal weighting of BNSL to LLM rather than 50/50}
  \label{tab:oracle}
  \begin{tabular}{p{0.26\textwidth} p{0.08\textwidth} p{0.08\textwidth} p{0.08\textwidth} p{0.08\textwidth} p{0.08\textwidth} p{0.16\textwidth}}
    \toprule
    Network & Nodes & 50/50 Fusion & {Oracle \newline Fusion} & Optimal BNSL Weight & Oracle \newline vs 50/50 & Oracle vs \newline best of LLM \newline and BNSL \\
    \midrule
alarm                     &    37 &   0.866 &   0.884 &  0.55 &  +0.018 &  +0.080 \\
asia                      &     8 &   1.000 &   1.000 &  0.25 &  +0.000 &  +0.158 \\
barley                    &    48 &   0.645 &   0.645 &  0.50 &  +0.000 &  +0.096 \\
child                     &    20 &   0.923 &   0.941 &  0.55 &  +0.018 &  +0.039 \\
constructionproductivity  &    18 &   0.826 &   0.919 &  0.90 &  +0.093 &  +0.054 \\
corical                   &    20 &   0.833 &   0.851 &  0.55 &  +0.018 &  +0.028 \\
diarrhoea                 &    28 &   0.750 &   0.752 &  0.60 &  +0.002 &  +0.120 \\
earthquake                &    40 &   0.602 &   0.611 &  0.55 &  +0.009 &  +0.027 \\
estuary                   &    30 &   0.721 &   0.762 &  0.30 &  +0.041 &  +0.147 \\
flood                     &    22 &   0.682 &   0.682 &  0.50 &  +0.000 &  +0.182 \\
hailfinder                &    56 &   0.717 &   0.735 &  0.55 &  +0.018 &  +0.049 \\
hepar2                    &    70 &   0.717 &   0.721 &  1.00 &  +0.004 &  +0.000 \\
insurance                 &    27 &   0.808 &   0.808 &  0.40 &  +0.000 &  +0.113 \\
kosterhavet               &    38 &   0.797 &   0.800 &  0.55 &  +0.003 &  +0.028 \\
macrophytes               &    15 &   0.947 &   0.947 &  0.50 &  +0.000 &  +0.043 \\
mildew                    &    35 &   0.787 &   0.804 &  0.25 &  +0.017 &  +0.101 \\
nuclearwaste              &    10 &   0.609 &   0.667 &  0.70 &  +0.058 &  +0.095 \\
permaBN                   &    14 &   0.766 &   0.766 &  0.50 &  +0.000 &  +0.085 \\
pneumonia                 &    62 &   0.702 &   0.702 &  0.50 &  +0.000 &  +0.065 \\
property                  &    27 &   0.952 &   0.984 &  0.55 &  +0.031 &  +0.158 \\
sachs                     &    11 &   0.727 &   0.788 &  0.55 &  +0.061 &  +0.182 \\
seismic                   &    10 &   0.692 &   0.769 &  0.20 &  +0.077 &  +0.049 \\
sports                    &     9 &   0.846 &   0.846 &  0.50 &  +0.000 &  +0.031 \\
urinary                   &    36 &   0.742 &   0.759 &  0.55 &  +0.017 &  +0.021 \\
vessel1                   &    16 &   0.789 &   0.909 &  0.65 &  +0.120 &  +0.187 \\
water                     &    32 &   0.322 &   0.421 &  0.25 &  +0.099 &  +0.024 \\
\midrule
AVERAGE                   &       &   0.760 &   0.787 &       &  +0.027 & +0.083 \\
    \bottomrule
  \end{tabular}
\end{table}

Table~\ref{tab:oracle} shows the further improvement that is gained if an optimum weighting of the BNSL and LLM contribution is used. The column headed "50/50 Fusion" repeats the F1 gain using 50/50 fusion reported in Table~\ref{tab:fusion}, which can be compared to the F1 obtained using an optimal weighting shown under "Oracle Fusion". The weighting given to the BNSL source which achieves this optimal F1 is shown in the "Optimal BNSL Weight" column. The penultimate column shows the improvement in F1 gained by using an optimal weighting compared to the default 50/50 weighting, and the final column reports the gain F1 using optimal weighting over the better of LLM and BNSL alone.

\section{BNSL Ensemble Effect: Per-network Detail}
\label{apdx:bnsl-ensemble-detail}

\begin{table}[!ht]
  \centering
  \caption{BNSL ensemble effect per network: F1 of the mean
    individual algorithm runs vs.\ the merged three-algorithm
    ensemble (10K samples).}
  \label{tab:bnsl-ensemble-detail}
  \begin{tabular}{p{0.26\textwidth} p{0.08\textwidth} lll}
    \toprule
    Network & Nodes & Mean Indiv. & Ensemble & Gain \\
    \midrule
alarm                     &    37 &   0.766 &   0.804 &  +0.038 \\
asia                      &     8 &   0.653 &   0.750 &  +0.097 \\
barley                    &    48 &   0.476 &   0.549 &  +0.073 \\
child                     &    20 &   0.791 &   0.902 &  +0.111 \\
constructionproductivity  &    18 &   0.767 &   0.865 &  +0.098 \\
corical                   &    20 &   0.644 &   0.727 &  +0.084 \\
diarrhoea                 &    28 &   0.533 &   0.632 &  +0.098 \\
earthquake                &    40 &   0.518 &   0.584 &  +0.066 \\
estuary                   &    30 &   0.435 &   0.435 &  $-$0.001 \\
flood                     &    22 &   0.515 &   0.474 &  $-$0.042 \\
hailfinder                &    56 &   0.643 &   0.686 &  +0.043 \\
hepar2                    &    70 &   0.559 &   0.721 &  +0.162 \\
insurance                 &    27 &   0.641 &   0.695 &  +0.053 \\
kosterhavet               &    38 &   0.693 &   0.772 &  +0.079 \\
macrophytes               &    15 &   0.839 &   0.905 &  +0.065 \\
mildew                    &    35 &   0.383 &   0.519 &  +0.136 \\
nuclearwaste              &    10 &   0.707 &   0.571 &  $-$0.136 \\
permaBN                   &    14 &   0.610 &   0.681 &  +0.071 \\
pneumonia                 &    62 &   0.509 &   0.637 &  +0.127 \\
property                  &    27 &   0.714 &   0.825 &  +0.112 \\
sachs                     &    11 &   0.293 &   0.606 &  +0.313 \\
seismic                   &    10 &   0.694 &   0.720 &  +0.026 \\
sports                    &     9 &   0.467 &   0.400 &  $-$0.067 \\
urinary                   &    36 &   0.555 &   0.738 &  +0.184 \\
vessel1                   &    16 &   0.780 &   0.722 &  $-$0.057 \\
water                     &    32 &   0.392 &   0.283 &  $-$0.109 \\
\midrule
AVERAGE                   &       &   0.599 &   0.662 &  +0.062 \\
    \bottomrule
  \end{tabular}
\end{table}

Table~\ref{tab:bnsl-ensemble-detail} shows the improvement which the baseline BNSL ensemble \emph{already achieves} over individual BNSL runs. The third column shows the mean F1 achieved averaging across all the individual BNSL runs performed by the FGES, Tabu-Stable and PC-Stable algorithms on the 10K row datasets. The "Ensemble" column shows the mean F1 achieved by using Algorithm~\ref{alg:merge_graphs} to create an ensemble PDG across these individual runs and then Algorithm~\ref{alg:to_dag_greedy} to create an ensemble DAG from this PDG. This averaging process improves F1 over the mean of individual runs in 20/26 networks, with a mean improvement in F1 of $+0.062$. The BNSL/LLM fusion process improves against this \emph{strengthened baseline.}

\section{Post-hoc Information Leakage Audit}
\label{apdx:leakage}

\subsection{Motivation}
\label{sub:leakage-motivation}

A concern raised during peer review was that some of the LLM performance observed in this study might arise from benchmark recognition or information leakage rather than causal reasoning. Two possible mechanisms are investigated here:

\begin{enumerate}
    \item Benchmark recognition, whereby the LLM identifies a known benchmark network and recalls information about its structure.
    \item Context-generation leakage, whereby variable aliases or descriptions generated during the preprocessing stage inadvertently reveal aspects of the reference graph.
\end{enumerate}

To investigate these possibilities, we conduct a post-hoc audit of all network context files used in the experiments.

\subsection{Audit Methodology}
\label{sub:leakage-methodology}

The audit was performed using Claude Opus 4.5 as an auditor. Claude Opus 4.5 was selected because the audit task requires open-ended reasoning about benchmark identifiability and semantic causal cues. The audit should therefore be viewed as an informed expert assessment rather than an objective measurement. For each benchmark network, the complete context supplied to the LLMs was provided to the auditor together with instructions to evaluate:

\begin{itemize}
    \item the likelihood that the benchmark could be identified from the supplied context information;
    \item the extent to which the supplied context information revealed causal structure.
\end{itemize}
according to the qualitative scores shown in Table~\ref{tab:leakage-scores}. 

\begin{table}[!ht]
  \centering
  \caption{Benchmark Identification and Causal Information Leakage Score Scales}
  \label{tab:leakage-scores}
  \begin{tabular}{p{0.07\textwidth} p{0.40\textwidth} p{0.40\textwidth}}
    \toprule
   Score  & Benchmark Identification & Causal Information Leakage \\
    \midrule
   0      & benchmark not identifiable                            & no causal information present \\
   3      & application domain recognisable but benchmark unknown & mild causal hints \\
   5      & plausible benchmark identification                    & several causal relationships inferable \\
   8      & strong benchmark match                                & most graph structure inferable \\
   10     & benchmark identifiable with high confidence           & graph largely recoverable from context alone \\
    \bottomrule
  \end{tabular}
\end{table}

The audit was performed with four settings:
\begin{enumerate}
    \item Minimal prompts using the obfuscated variable names used in the experiments.
    \item Standard prompts using the obfuscated variable names used in the experiments.
    \item Minimal prompts using the original benchmark variable names.
    \item Standard prompts using the original benchmark variable names.
\end{enumerate}
The auditor was additionally asked to justify its assessment by identifying specific pieces of information that contributed to the assigned score.

\subsection{Audit Results}
\label{sub:leakage-results}

\begin{table}[!ht]
  \centering
  \caption{Scores out of 10 for benchmark identification estimated by Claude Opus 4.5.}
  \label{tab:leakage-benchmark}
  \begin{tabular}{p{0.14\textwidth} p{0.08\textwidth} p{0.11\textwidth} p{0.11\textwidth} p{0.11\textwidth} p{0.11\textwidth} p{0.14\textwidth} }
    \toprule
            &       
            & \multicolumn{2}{c}{Obfuscated names}
            & \multicolumn{2}{c}{Benchmark names} 
            & \multirow{2}{*}{\shortstack[l]{Fusion gain\\in F1 over\\better of\\LLM/BNSL}}\\
    \textcolor{white}{blank line} \newline Network  
    & \textcolor{white}{blank} \newline Nodes 
    & Minimal \newline prompt
    & Standard \newline prompt
    & Minimal \newline prompt
    & Standard \newline prompt & \\
    \midrule
    alarm        &    37 &   9 &   9 &  10 &  10 & +0.062 \\
    asia         &     8 &   7 &   8 &  10 &  10 & +0.158 \\
    barley       &    48 &   8 &   8 &   4 &   8 & +0.096 \\
    child        &    20 &   9 &   8 &   9 &   9 & +0.021 \\
    construct.   &    18 &   2 &   6 &   5 &   7 & -0.039 \\
    corical      &    20 &   6 &   6 &   6 &   8 & +0.010 \\
    diarrhoea    &    28 &   7 &   7 &   6 &   4 & +0.118 \\
    earthquake   &    40 &   2 &   4 &   6 &   7 & +0.018 \\
    estuary      &    30 &   7 &   8 &   7 &   8 & +0.106 \\
    flood        &    22 &   2 &   6 &   3 &   6 & +0.182 \\
    hailfinder   &    56 &   9 &   9 &   9 &   9 & +0.031 \\
    hepar2       &    70 &   7 &   8 &   7 &   9 & -0.004 \\
    insurance    &    27 &   8 &   9 &   9 &   9 & +0.113 \\
    kosterhavet  &    38 &   3 &   6 &   8 &   7 & +0.025 \\
    macrophytes  &    15 &   3 &   7 &   7 &   7 & +0.043 \\
    mildew       &    35 &   2 &   2 &   7 &   9 & +0.084 \\
    nuclearwaste &    10 &   3 &   4 &   3 &   7 & +0.037 \\
    permaBN      &    11 &   2 &   2 &   2 &   2 & +0.085 \\
    pneumonia    &    62 &   7 &   7 &   8 &   7 & +0.065 \\
    property     &    27 &   2 &   4 &   2 &   7 & +0.127 \\
    sachs        &    11 &   9 &   9 &   9 &   9 & +0.121 \\
    seismic      &    10 &   3 &   7 &   4 &   7 & -0.028 \\
    sports       &     9 &   2 &   3 &   2 &   6 & +0.031 \\
    urinary      &    36 &   3 &   2 &   3 &   3 & +0.003 \\
    vessel1      &    16 &   6 &   7 &   8 &   8 & +0.067 \\
    water        &    32 &   4 &   7 &   8 &   8 & -0.075 \\
    \bottomrule
  \end{tabular}
\end{table}

Table~\ref{tab:leakage-benchmark} shows the benchmark identification score estimated by Claude for each network. Several well-known networks such as alarm, asia, child, hailfinder, insurance and sachs are judged to be very susceptible to being identified even with the obfuscated variable names used in the experiments. However, many lesser-known networks including earthquake, mildew, permaBN, property and urinary seem more resistant to benchmark identification. Looking at the narrative report provided by Claude for each network shows that the total number of variables, and numbers of variables of particular type, as well as the variable names themselves may be used to identify the network. For example, it identified the alarm network because it had 37 variables, groups of which related to ventricular or heart rate measurements. For the well-known benchmarks it even provided a mapping between obfuscated and benchmark variable names.  The variable name obfuscation approach used in the experiments reduces benchmark identification in some cases, for example mildew and earthquake, especially when minimal prompts were used. Comparing the benchmark identification scores using obfuscated names with the fusion gain in F1 (shown in the final column in the Table) shows a weak, statistically non-significant Spearman correlation of $\rho=0.14$ using minimal prompts, and $\rho=0.18$ with standard prompts.

Table~\ref{tab:leakage-causal} shows the estimates for causal information leakage through variable names, and with standard prompting, variable descriptions, types and states. For many networks this is judged to be relatively mild, with the highest values for corical and vessel1. Scores were generally very similar across the two prompt levels and variable naming regimes for each network. This suggests that the additional information provided in standard prompts was not adding much causal information beyond that provided by variable names, and that the obfuscated and benchmark variable names provided similar amounts of causal information.

Some examples of causal leakage from variable name highlighted by Claude were \say{vaccine\_induced\_coagulapathy} and \say{infection\_portal\_thrombosis} in the corical network, or a causal relationship between \say{unstable\_cargo\_loading} and \say{stability\_loss} in the vessel1 network. Other, perhaps less obvious examples, revolved around including words such \say{outcome} or \say{score} which tend to be linked to leaf nodes, and \say{baseline} or \say{input} which tend to suggest upstream causes. Despite these examples, analysis showed that the estimated causal leakage score was weakly correlated with the F1 fusion gain - for obfuscated names, it was $\rho = -0.19$ with minimal prompts and $\rho = -0.09$ with standard prompts. We note too that mean fusion gain with standard prompts was 0.730 compared to 0.696, so that the likely additional causal information provided in descriptions did not have a big influence on fusion performance.

\begin{table}[!ht]
  \centering
  \caption{Scores out of 10 for causal information leakage estimated by Claude Opus 4.5.}
  \label{tab:leakage-causal}
  \begin{tabular}{p{0.14\textwidth} p{0.08\textwidth} p{0.11\textwidth} p{0.11\textwidth} p{0.11\textwidth} p{0.11\textwidth} p{0.14\textwidth} }
    \toprule
            &       
            & \multicolumn{2}{c}{Obfuscated names}
            & \multicolumn{2}{c}{Benchmark names} 
            & \multirow{2}{*}{\shortstack[l]{Fusion gain\\in F1 over\\better of\\LLM/BNSL}}\\
    \textcolor{white}{blank line} \newline Network  
    & \textcolor{white}{blank} \newline Nodes 
    & Minimal \newline prompt
    & Standard \newline prompt
    & Minimal \newline prompt
    & Standard \newline prompt & \\
    \midrule
    alarm        &    37 &   4 &   4 &   4 &   4 & +0.062 \\
    asia         &     8 &   4 &   6 &   4 &   7 & +0.096 \\
    child        &    20 &   6 &   4 &   4 &   4 & +0.021 \\
    construct.   &    18 &   4 &   4 &   4 &   4 & -0.039 \\
    corical      &    20 &   7 &   7 &   7 &   7 & +0.010 \\
    diarrhoea    &    28 &   4 &   4 &   4 &   3 & +0.118 \\
    earthquake   &    40 &   7 &   6 &   4 &   6 & +0.018 \\
    estuary      &    30 &   6 &   6 &   6 &   6 & +0.106 \\
    flood        &    22 &   4 &   4 &   4 &   4 & +0.182 \\
    hailfinder   &    56 &   6 &   6 &   4 &   4 & +0.031 \\
    hepar2       &    70 &   4 &   4 &   4 &   6 & -0.004 \\
    insurance    &    27 &   4 &   4 &   4 &   4 & +0.113 \\
    kosterhavet  &    38 &   4 &   4 &   4 &   4 & +0.025 \\
    macrophytes  &    15 &   4 &   5 &   6 &   5 & +0.043 \\
    mildew       &    35 &   6 &   6 &   4 &   6 & +0.084 \\
    nuclearwaste &    10 &   4 &   5 &   4 &   4 & +0.037 \\
    permaBN      &    11 &   4 &   4 &   5 &   4 & +0.085 \\
    pneumonia    &    62 &   6 &   4 &   6 &   6 & +0.065 \\
    property     &    27 &   6 &   6 &   6 &   6 & +0.127 \\
    sachs        &    11 &   4 &   4 &   2 &   4 & +0.121 \\
    seismic      &    10 &   6 &   6 &   6 &   6 & -0.028 \\
    sports       &     9 &   4 &   4 &   4 &   4 & +0.031 \\
    urinary      &    36 &   4 &   4 &   4 &   4 & +0.003 \\
    vessel1      &    16 &   7 &   7 &   7 &   7 & +0.067 \\
    water        &    32 &   6 &   6 &   4 &   5 & -0.075 \\
    \bottomrule
  \end{tabular}
\end{table}

\subsection{Summary and Future Work}
\label{tab:leakage-future}

This post-hoc audit cannot conclusively determine whether benchmark memorisation and causal information leakage contributed to the observed results. The scores were generated by an LLM and therefore represent informed qualitative judgements rather than objective measurements.

Nevertheless, the audit provides three pieces of evidence against benchmark-recognition or context-generation leakage being the primary explanation for the observed fusion gains:
\begin{enumerate}
    \item variable-name obfuscation substantially reduced benchmark identifiability for many networks;
    \item benchmark-identification scores were not associated with fusion improvement;
    \item estimated causal information leakage was not associated with fusion improvement.
\end{enumerate}
This audit also provided some useful insights as to approaches which could quantify and reduce the effects of memorisation further:
\begin{itemize}
    \item remove some variables, including within specific groups of variables to hinder benchmark identification
    \item obfuscate variable names with specific attention to removing all causal clues;
    \item use lesser-known, or ideally theoretically-grounded novel, benchmark networks
\end{itemize}
\end{document}